\documentclass[11pt,a4paper]{article}
\usepackage{emnlp-ijcnlp-2019}
\usepackage{times}
\usepackage{latexsym}

\usepackage{url}

\aclfinalcopy 

\usepackage{array}
\newcolumntype{P}[1]{>{\centering\arraybackslash}p{#1}}

\usepackage{amsmath}
\usepackage{amsfonts}
\usepackage{graphicx}
\usepackage{booktabs}
\usepackage{enumerate}
\usepackage{multirow}

\usepackage{arydshln}
\usepackage[english]{babel}
\usepackage[autostyle, english=american]{csquotes}
\MakeOuterQuote{"}
\usepackage{subcaption}
\usepackage{soul}
\usepackage{mwe}

\title{Deleter: Leveraging BERT to Perform Unsupervised Successive Text Compression} 

\author{
    Tong Niu $\;\;\;\;\;$ Caiming Xiong $\;\;\;\;\;$ Richard Socher \\
    Salesforce Research \\
  {\tt \{tniu, cxiong, rsocher\}@salesforce.com}
}

\date{}

\begin{document}
\maketitle

\begin{abstract}
Text compression has diverse applications such as Summarization, Reading Comprehension and Text Editing. However, almost all existing approaches require either hand-crafted features, syntactic labels or parallel data. Even for one that achieves this task in an unsupervised setting, its architecture necessitates a task-specific autoencoder. Moreover, these models only generate one compressed sentence for each source input, so that adapting to different style requirements (e.g. length) for the final output usually implies retraining the model from scratch. In this work, we propose a fully unsupervised model, \textsc{Deleter}, that is able to discover an "\textit{optimal deletion path}" for an arbitrary sentence, where each intermediate sequence along the path is a coherent subsequence of the previous one. This approach relies exclusively on a pretrained bidirectional language model (BERT) to score each candidate deletion based on the average Perplexity of the resulting sentence and performs progressive greedy lookahead search to select the best deletion for each step. We apply \textsc{Deleter} to the task of extractive Sentence Compression, and found that our model is competitive with state-of-the-art supervised models trained on $1.02$ million in-domain examples with similar compression ratio. Qualitative analysis, as well as automatic and human evaluations both verify that our model produces high-quality compression.
\end{abstract}
\begin{table*}[t]
  \centering
    \begin{tabular}{r|cc}
    \multicolumn{1}{c|}{Sentence} & \multicolumn{1}{c}{Deleted Tokens} & \multicolumn{1}{c}{AvgPPL} \\
    \midrule
    i think america is still a fairly crowded country by the way . & -     & 5.72 \\
    i think america is a fairly crowded country by the way . & still & 5.97 \\
    i think america is a crowded country by the way . & fairly & 7.19 \\
    i think america is a crowded country . & by the way & 5.96 \\
    i think america is a country . & crowded & 7.09 \\
    america is a country . & i think & 8.15 \\
    america . & is a country & 7.55 \\
    \end{tabular}%
  \caption{The optimal deletion path our model finds (without applying a termination condition) for the sentence "\textit{i think america is still a fairly crowded country by the way .}"}
  \label{tab:path1}%
\end{table*}%

\section{Introduction}
\label{sec:introduction}
Text compression can be applied to various tasks such as Summarization, Text Editing and even Data Augmentation where the compressed text can be employed as additional training examples. However, almost all existing approaches require either parallel data, hand-crafted rules, or extra syntactic information such as dependency labels or part-of-speech tag features trees~\citep{mcdonald2006discriminative,filippova2008dependency,zhao2018language}. Even for one model that achieves this task in an unsupervised setting, its architecture necessitates a task-specific "sequence-to-sequence-to-sequence" autoencoder~\citep{baziotis2019seq}. Moreover, these models only generate one compressed sentence for each source input, making adaption to different style requirements difficult.

In this work, we propose a fully unsupervised model, \textsc{Deleter}, that is able to discover an "\textit{optimal deletion path}" for a sentence, where each intermediate sequence along the path is a coherent subsequence of the previous one. \textsc{Deleter} relies exclusively on a pretrained bidirectional language model, BERT~\citep{devlin2018bert}, to score each candidate deletion based on the average Perplexity of the resulting sentence and performs progressive greedy look-ahead search to select the best deletion(s) for each step.\footnote{We will release all our code and model outputs soon.} Table~\ref{tab:path1} shows a real example of how the \textsc{Deleter} \textbf{gradually} turns "\textit{i think america is still a fairly crowded country by the way .}" into "\textit{america is a country .}", where each intermediate sequence between them is a completely grammatical sentence. Interestingly, the model can also be used to delete extra words in a sentence, such as turning "\textit{i \textbf{work work} at a company .}" into "\textit{i \textbf{work} at a company .}"

As shown in the table, unlike previous pure-black-box approaches, our model not only provides the final compression, but also exposes the full deletion path, making the compression process more comprehensible and controllable, so that it is easy to freeze certain key words from the original sentence or enforce a certain minimum/maximum compression ratio.

We apply \textsc{Deleter} to the task of extractive Sentence Compression, and found that our model is competitive with state-of-the-art supervised models trained on $1.02$ million in-domain examples. Qualitative analysis, as well as automatic and human evaluations, both verify that our model produces high-quality compression.

\section{The \textsc{Deleter} Model}
\label{sec:model}
Our \textsc{Deleter} employs \textit{progressive lookahead greedy search} based on a pretrained BERT, which is used to assign a negative log likelihood for each token in a sentence to derive a score for each intermediate sentence. The algorithm aims to minimize the average score along the path to ensure that each intermediate sentence is grammatical.

\vspace{-4pt}
\paragraph{Task Formulation}
Given a sentence, there are finite possible tokens we could keep deleting until running out of tokens. In our setting, we do not go back and add tokens to an intermediate sentence. We thus formulate the task of Sentence Compression as finding an \textit{optimal deletion path} along a Directed Acyclic Graph.
Each node in this graph is either the original sentence itself (i.e., the root node, which has no incoming edges) or a subsequence of it. Each node has outgoing edges pointing to each sentence resulting from a legal deletion of the current node (note that our model allows deletion of multiple tokens in one step). Each edge is also associated with a cost, which is the \textit{Average Perplexity Score} (AvgPPL) of all tokens in the sentence as assigned by BERT. We consider a path optimal if the average score of all nodes along the path is minimized.


\vspace{-4pt}
\paragraph{Sentence Scoring}
The \textsc{Deleter} assigns each sentence in the graph an AvgPPL score as shown in the equation below. Lower score indicates more syntactically coherent sentence.
\begin{align*}
    AvgPPL &= \exp \Bigg( - \sum_{i=1}^m \log T_i - \sum_{i=1}^n \log D_i \Bigg)
\label{eq:avgppl}
\end{align*}
$\{T_i\}$ are tokens present in the current sentence and $\{D_i\}$ refer to the collection of all deleted tokens. Note that each $\log D_i$ is obtained by scoring the corresponding token in the \textbf{original sentence}. This second term is crucial to the algorithm because otherwise the model would prefer to delete rarer tokens, whose unigram probabilities are lower (i.e., of higher negative log likelihood).


\begin{figure}[t]
\centering
\includegraphics[width=0.47\textwidth]{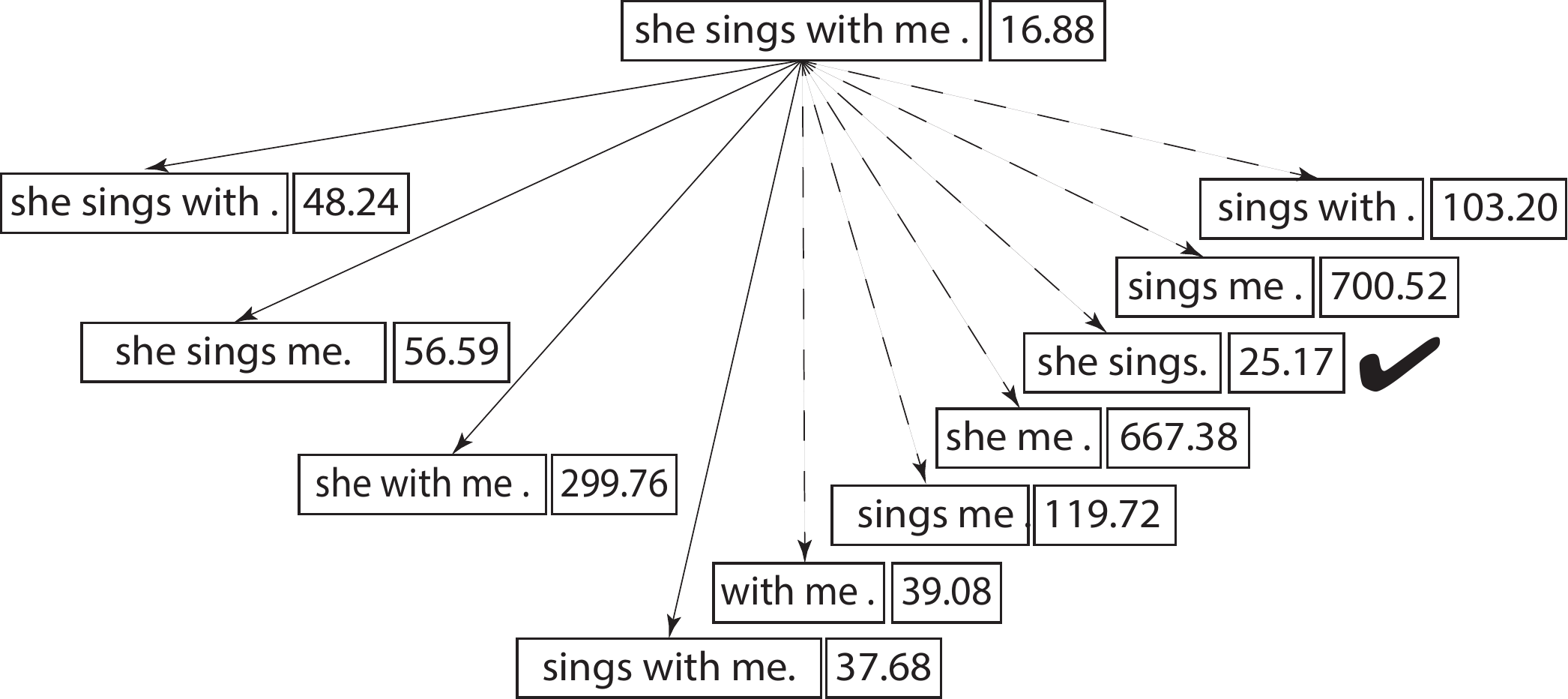}
\vspace{-5pt}
\caption{How progressive lookahead greedy search probes for the next best deletion. Dotted arrow lines correspond to the second iteration of probing because the first iteration (non-dotted arrow lines) does not locate a good deletion candidate. The model then chooses the sentence with the lowest AvgPPL score (the one with a check mark on the right).}
\label{fig:lookahead}
\vspace{-8pt}
\end{figure}

\vspace{-6pt}
\paragraph{Progressive Lookahead Greedy Search}
\textsc{Deleter} allows deleting multiple tokens at a time, but naively probing for deletions consisting of multiple tokens will result in computation workload scaling with $O(m^k)$ , where $m$ is the the number of tokens in the current intermediate sentence and $k$ is the maximum deletion length for each step. We thus adopt a progressive approach to help reduce the search space. Figure~\ref{fig:lookahead} shows how a real probing step looks like for the sentence "\textit{she sings with me .}". For each step, we first probe for deletions consisting of only one token; if no candidate obtains a score below a threshold, we increase the lookahead length to $2$, and so forth. A candidate is above this threshold $T$ if:
\begin{align*}
    \frac{AvgPPL_{i + 1}}{AvgPPL_{i}} &> 1 + \alpha \log (L_{root})
\end{align*}
where $\alpha$ (set to $0.04$ in our experiments) is a hyperparameter controlling the extent of increase on AvgPPL for each step. The lower it is, the better the compression quality (with the price of being slower). $L_{root}$ is the number of tokens in the original sentence. We include this term because shorter sentences tend to have lower percentage of increase on AvgPPL, in which case we want to lower the threshold. Note that during inference, this threshold also functions as the termination condition in case we are only allowed to select one sentence from the deletion path (e.g. for Sentence Compression).

During probing, we also want to slightly discourage the model from deleting multiple tokens at a time, since we want it to traverse the deletion path as gradually as possible to have more intermediate compressions. We thus further multiply AvgPPL by $L_s^{\beta}$ when probing for the next deletion, where $L_s$ is the length of the current sentence and $\beta$ (set to $0.04$ in our experiments) indicates how "gradual" we want the \textsc{Deleter} to proceed.





\section{Experimental Setup}
\label{sec:experimental setup}
\paragraph{\textsc{Deleter} Model Details}
We use a pretrained BERT uncased model implemented by HuggingFace.\footnote{
https://github.com/huggingface/pytorch-pretrained-BERT
} To score any token in a sentence, we use the special token [mask] to mask out the target token, and then prepend a [CLS] and append a [SEP], which function as [START] and [END] tokens. We use both tokens to help the model evaluate whether the current first or last token can function as a sentence start or end. This is also the main reason that we did not choose other competitive language models such as GPT-2~\citep{radford2019language}: because these models are not bidirectional, they will be much less sensitive to deletions from the very end of a sentence. However, note that since BERT is not intended to be a language model,\footnote{Though~\citet{wang2019bert} found that BERT can be formulated as a Markov Random Field language model.} we mask one token at a time to obtain the negative log probability of each. The maximum lookahead length in our work is set to $3$ to facilitate deletion of phrases which contain multiple tokens, such as "\textit{by the way}" in Table~\ref{tab:path1}.

\paragraph{Datasets}
We experiment on two datasets. Google sentence compression dataset~\citep{filippova2013overcoming} contains $200,000$ training examples, information of compression labels, part-of-speech (POS) tags, and dependency parsing information for each sentence.\footnote{
https://github.com/google-research-datasets/sentence-compression
} Following~\citet{filippova2015fast}, we used the first $1,000$ sentences of evaluation set as our test set. 
The Gigaword dataset~\citep{napoles2012annotated} with $1.02$ million examples, where the first $200$ are labeled for extractive Sentence Compression by two annotators~\citep{zhao2018language}.\footnote{https://github.com/code4conference/Data}


\vspace{-1pt}
\paragraph{Automatic and Human Evaluation}
Following~\citet{wang2017can}, we use the ground truth compressed sentences to compute F1 scores. 
Although F1 can be indicative of how well a compression model performs, we note that their could be multiple viable compressions for the same sentence, which single-reference ground-truth cannot cover~\citep{handler2019human}. Thus to faithfully evaluate the quality of the compressions generated by our model, we follow~\citet{zhao2018language} and conducted human studies on the Gigaword dataset with Amazon Mechanical Turk (MTurk).\footnote{
https://www.mturk.com/
}

Specifically, we sample $150$ examples from the test set, and put our model output side-by-side with the two compressions created by the two annotators.\footnote{We did not compare with the best-performing model in~\citet{zhao2018language} because we did not succeed in running their published code on Github:
https://github.com/code4conference/code4sc
} The three compressions were randomly shuffled to anonymize model identity. We hire annotators who have an approval rate of at least $98\%$ and $10,000$ approved HITs. Following~\citep{zhao2018language}, we employed Readability and Informativeness as criteria on a five-point Likert scale.

\section{Results and Analysis}
\label{sec:results}
\paragraph{Automatic Evaluation}
We report F1 scores on both Google and Gigaword dataset. Since each of the $200$ sentences from Giga test set has two references (by Annotator 1 and 2, respectively), we report two F1's following~\citet{zhao2018language}. As shown in Table~\ref{tab:f1-giga}, our model is competitive (esp. for Annotator \#2) with state-of-the-art supervised models trained on $1.02$ million examples with similar compression ratio.

\begin{table}[t]
\vspace{-7pt}
\small
  \centering
    \begin{tabular}{r|ccc}
          & F1 (\#1) & F1 (\#2) & CR \\
    \midrule
    Seq2seq with attention & 54.9  & 58.6  & 0.53 \\
    Dependency tree+ILP & 58.0    & 61.0    & 0.55 \\
    LSTMs+pseudo label & 60.3  & 64.1  & 0.51 \\
    Evaluator-LM & 64.5  & 66.9  & 0.50 \\
    Evaluator-SLM & \textbf{65.0} & \textbf{68.2} & 0.51 \\
    Deleter & 54.0    & 62.6  & 0.56 \\
    \end{tabular}%
  \caption{F1 results for both Annotator \#1 and \#2 on Gigaword dataset. CR stands for Compression Ratio. Best results are boldfaced.}
  \label{tab:f1-giga}%
  \vspace{-7pt}
\end{table}%

We also report F1 results on the Google dataset (Table~\ref{tab:human}). We can see that for this dataset our F1 is still far away from state-of-the-art. We reason that this is partly because the ground-truth compressions in Google dataset are based on news headlines (later edited by human) rather than compressions generated directly by human beings.\footnote{For example, in Google dataset the sentence "\textit{mcdonalds is teaching their employees to say 'no' to fast food.}" is compressed to be "\textit{mcdonalds is teaching to say to fast food}.", which does not make sense.}

\begin{table}[t]
\small
  \centering
    \begin{tabular}{r|cc}
    Seq2seq with attention & F1    & CR \\
    \midrule
    LSTM  & 71.7  & 0.34 \\
    LSTMs & 82.0    & 0.38 \\
    Evaluation-LM & 84.8  & 0.4 \\
    EValuation-SLM & \textbf{85.0}    & 0.41 \\
    Deleter & 50.0    & 0.39 \\
    \end{tabular}%
  \caption{F1 results on the Google Compression dataset. CR stands for Compression Ratio. Best results are boldfaced.}
  \label{tab:f1-google}%
  \vspace{-10pt}
\end{table}%

\paragraph{Human Evaluation}
As mentioned in Section~\ref{sec:experimental setup}, we conducted human evaluation on Gigaword. As shown in Table~\ref{tab:human}, the Readability of our model is close to that of Annotator \#1, while there is a larger gap on Informativeness between the two. This is expected because our model is syntax-aware rather than semantic-aware. For example, the \textsc{Deleter} would readily delete negation words (e.g., from "i do not love NLP" to "i do love NLP"), which usually causes abrupt change in meaning. We leave enforcing semantic consistency by leveraging Neural Language Inference datasets~\citep{williams2017broad} as future work. It is also interesting to note that there is a larger gap between Annotator \#1 and \#2 than between Annotator \#1 and our model, indicating that the quality of compression is a fairly subject matter.

\begin{table}[t]
\small
  \centering
    \begin{tabular}{r|cc}
          & \multicolumn{1}{l}{Readability} & \multicolumn{1}{l}{Informativeness} \\
    \midrule
    Our Model & 2.88  & 2.95 \\
    Annotator \#1 & 3.01  & 3.21 \\
    Annotator \#2 & \textbf{3.31}  & \textbf{3.56} \\
    \end{tabular}%
  \caption{Human evaluation results on the Gigaword dataset. Best results are boldfaced. Note that we are directly \textbf{comparing with human-generated compressions}. All differences (within same column) are stat. significant with $p < 0.05$.}
  \label{tab:human}%
\end{table}%

\section{Related Work}
\label{sec:related work}
Sentence Compression task has been investigated by various previous work~\citep{jing2000sentence,knight2000statistics,clarke2006constraint,mcdonald2006discriminative,clarke2008global,filippova2008dependency,berg2011jointly}, where the more recent work tend to adopt a neural approach~\citep{filippova2015fast,wang2017can,kamigaito2018higher,cifka2018eval}. Our model is also neural-based since it leverages a neural language model. Our work differs from previous work in that it does not require any syntactic information such as POS tags or dependency parse tree, which is an advantage especially for low-resource language where training data for tagger or parser is scarce. We note that~\citet{baziotis2019seq} also built an unsupervised model for abstractive sentence compression. They trained a "sequence-to-sequence-to-sequence" autoencoder to first compress the original sentence, and then reconstruct it. Similar to our AvgPPL,~\citet{zhao2018language} also employed average Perplexity (though without our length correction terms) as the reward to a policy network trained with reinforcement learning. 

Another characteristic of our model is that we obtain a sequence of sentences which are all valid compressions of the original one, while other models usually generate only one compression.

\vspace{-1pt}
\section{Conclusion}
\label{sec:conclusion}
We introduced \textsc{Deleter} which finds an optimal deletion path of a sentence where each node along the path is a grammatical subsequence. We applied this model to two Sentence Compression datasets and found that they are comparable with state-of-the-art supervised models. Note that Sentence Compression has not explored the full power of this model since it only selects one sentence as the output. We plan to apply \textsc{Deleter} to tasks such as Data Augmentation, where the training data of each epoch is dynamically assembled by randomly sampling an intermediate sentence along the deletion path of each sentence. This approach can also be leveraged as a way to generate adversarial examples~\citep{niu2018adversarial} because deleting a few tokens usually preserve semantics of the original sentence. 

\bibliography{references}
\bibliographystyle{acl_natbib}

\end{document}